%% file: main.tex
\documentclass[10pt,twocolumn,letterpaper]{article}

\usepackage{iccv}              

\usepackage{times}
\usepackage{epsfig}
\usepackage{graphicx}
\usepackage{amsmath}
\usepackage{amssymb}

\usepackage{graphicx}
\usepackage{booktabs}
\usepackage[symbol]{footmisc}

\usepackage{hyperref}

\def\paperID{4826} 
\def\confName{ICCV}
\def\confYear{2025}

\title{MinCD-PnP: Learning 2D-3D Correspondences with Approximate Blind PnP}

\author{
Pei An$^{1}$\footnotemark{}$\hspace{4pt}\hspace{8pt}$Jiaqi Yang$^{2}$\footnotemark[1]$\hspace{4pt}\hspace{8pt}$Muyao Peng$^{1}\hspace{8pt}$You Yang$^{1}$\footnotemark{}$\hspace{8pt}$Qiong Liu$^{1}\hspace{8pt}$Xiaolin Wu$^{3}$\hspace{8pt}Liangliang Nan$^{4}$\\
$^{1}$Huazhong University of Science and Technology, China\\
$^{2}$Northwestern Polytechnical University, China\\$^{3}$McMaster University, Canada\hspace{10pt}$^{4}$Delft University of Technology, Netherlands\\
}

\begin{document}
\maketitle

\footnotetext[1]{Equal contribution.}
\footnotetext[2]{Corresponding author: yangyou@hust.edu.cn.}

\begin{abstract}

Image-to-point-cloud (I2P) registration is a fundamental problem in computer vision, focusing on establishing 2D-3D correspondences between an image and a point cloud. The differential perspective-n-point (PnP) has been widely used to supervise I2P registration networks by enforcing the projective constraints on 2D-3D correspondences. However, differential PnP is highly sensitive to noise and outliers in the predicted correspondences. This issue hinders the effectiveness of correspondence learning. Inspired by the robustness of blind PnP against noise and outliers in correspondences, we propose an approximated blind PnP based correspondence learning approach. To mitigate the high computational cost of blind PnP, we simplify blind PnP to an amenable task of minimizing Chamfer distance between learned 2D and 3D keypoints, called MinCD-PnP. To effectively solve MinCD-PnP, we design a lightweight multi-task learning module, named as MinCD-Net, which can be easily integrated into the existing I2P registration architectures. Extensive experiments on 7-Scenes, RGBD-V2, ScanNet, and self-collected datasets demonstrate that MinCD-Net outperforms state-of-the-art methods and achieves a higher inlier ratio (IR) and registration recall (RR) in both cross-scene and cross-dataset settings.


\end{abstract}
\thispagestyle{plain} 

\section{Introduction}

Image-to-point-cloud (I2P) registration \cite{2d3d-match} is a fundamental task in computer vision \cite{an-new-add-1}, which aims to establish 2D-3D correspondences between images and point clouds \cite{p2-net}. These correspondences are used to estimate the six-degree-of-freedom (6 DoF) camera pose with perspective-n-point (PnP) algorithm \cite{epnp}, enabling I2P registration by aligning captured images with point clouds. Thus, I2P registration is widely used in visual localization, navigation, visual odometry, 3D reconstruction, and so on \cite{an-survey-1, ep2p-loc, visual_route, Pollefeys-1, Pollefeys-2}.

\begin{figure}[t]
	\centering
		\includegraphics[width=1.0\linewidth]{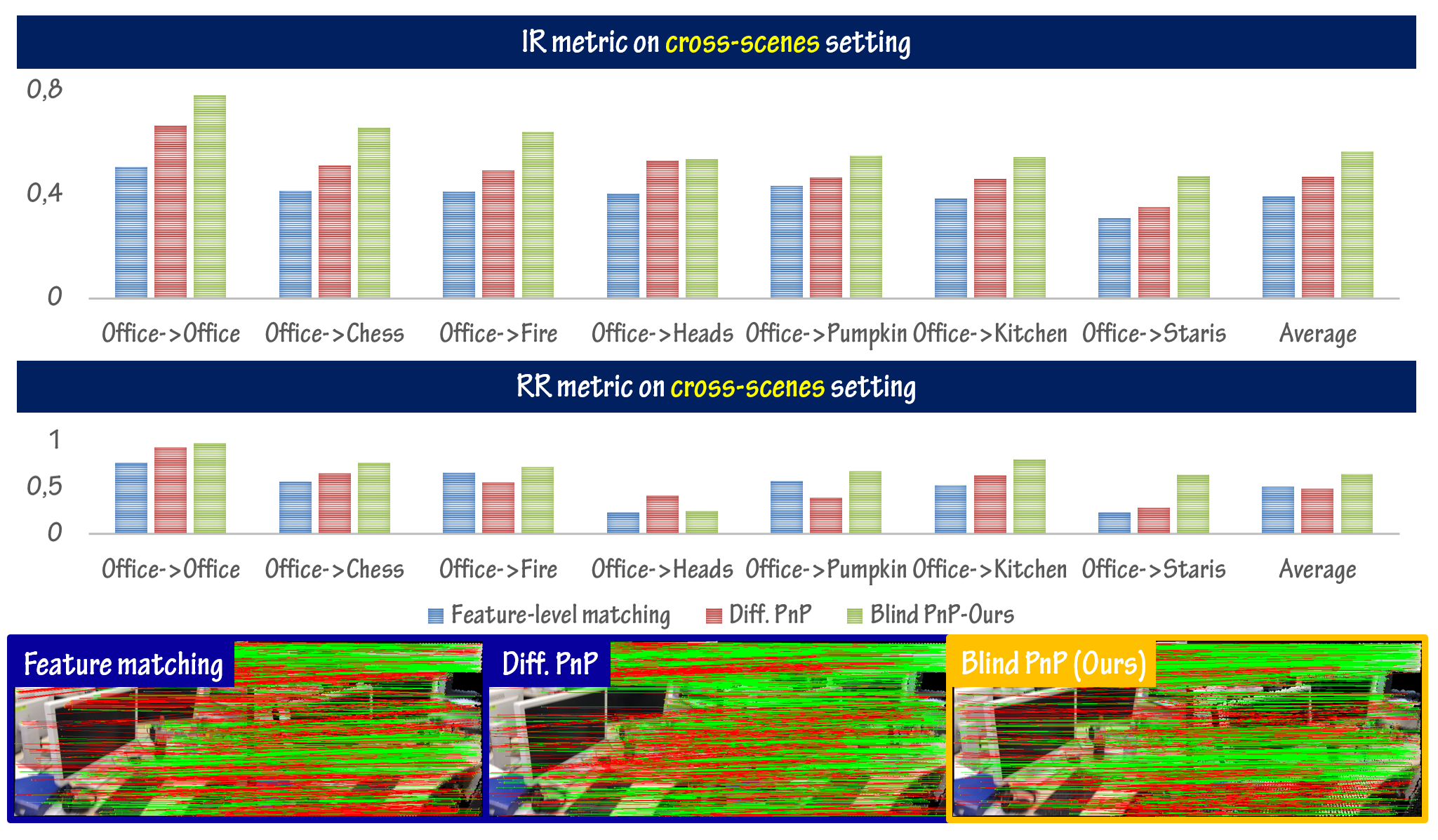}		
        \caption{To overcome the limitation of \textbf{feature-level matching}, \textbf{differential PnP} employs the projective constraints of 2D-3D correspondences but is highly sensitive to correspondence quality. In this paper, we incorporate \textbf{blind PnP} to enhance I2P registration, and achieve the salient improvement compared to other methods.}
	\label{fig:into}
\end{figure}

Learning-based approaches have gained significant attention in I2P registration \cite{an-survey-1, p2-net}. Deep neural networks (DNNs) help bridge  the modality gap between images and point clouds \cite{an-new-add-1,2d3d-matr}. They estimate 2D-3D correspondences by {pixel-to-point feature-level matching} (i.e., comparing the feature distance between each 2D pixel and each 3D point) \cite{2d3d-match}. However, feature-level matching struggles to remove outliers, because it ignores the projective constraints on 2D-3D correspondences, as shown in Fig. \ref{fig:into}.

In order to utilize the constraints of projective geometry in learning 2D-3D correspondences, the mainstream technique leverages the differential perspective-n-point (PnP) \cite{blind-pnp, Epro-PnP, diff_reg_match}. The objective is to refine camera pose estimation via differential PnP, thereby improving the accuracy of global projective correspondences. However, differential PnP is highly sensitive to noise and outliers in the predicted correspondences \cite{pnp-sensitive}. This issue makes the estimated camera pose unreliable, thus hindering the effectiveness of differential PnP on correspondence learning.



To overcome the limitations of differential PnP, inspired by 
the robustness of blind PnP against noise and outliers in correspondences \cite{go-blind-pnp}, we propose an approximate blind PnP based 2D-3D correspondence learning approach. Since blind PnP is computationally expensive \cite{go-blind-pnp}, we 
reformulate the original blind \textbf{PnP} as a task of \textbf{min}imizing \textbf{C}hamfer \textbf{d}istance between the learned 2D and 3D keypoints, called \textbf{MinCD-PnP} in the sequel.
MinCD-PnP ensures the feasibility of learning correspondence with blind PnP, and retains the robustness to noise and outliers in correspondences. 
To effectively solve MinCD-PnP, we propose a lightweight multi-task learning module, denoted by 
\textbf{MinCD-Net}.  Operationally, MinCD-Net can be easily integrated into the existing I2P registration architectures and jointly optimized in an end-to-end manner. 

Extensive experiments on the 7-Scenes \cite{scene-7-dataset}, RGBD-V2 \cite{RGBD-dataset}, ScanNet \cite{scan-net}, and self-collected datasets show that MinCD-Net achieves a higher inlier ratio (IR) and registration recall (RR) than state-of-the-art methods in both cross-scene and cross-dataset settings. Source code is released\footnote{https://github.com/anpei96/mincd-pnp-demo}. Our core contributions are: 



\begin{itemize}
    \item We introduce a task, MinCD-PnP, which simplifies blind PnP to a more amenable problem of minimizing Chamfer distance between learned 2D and 3D keypoints.
    \item To effectively solve MinCD-PnP, we design a lightweight multi-task learning module, MinCD-Net. It can be easily integrated into existing I2P registration architectures. 
    \item MinCD-Net outperforms existing methods in both cross-scene and cross-dataset settings on five public datasets. 
\end{itemize}

\begin{figure*}[t]
	\centering
		\includegraphics[width=1.0\linewidth]{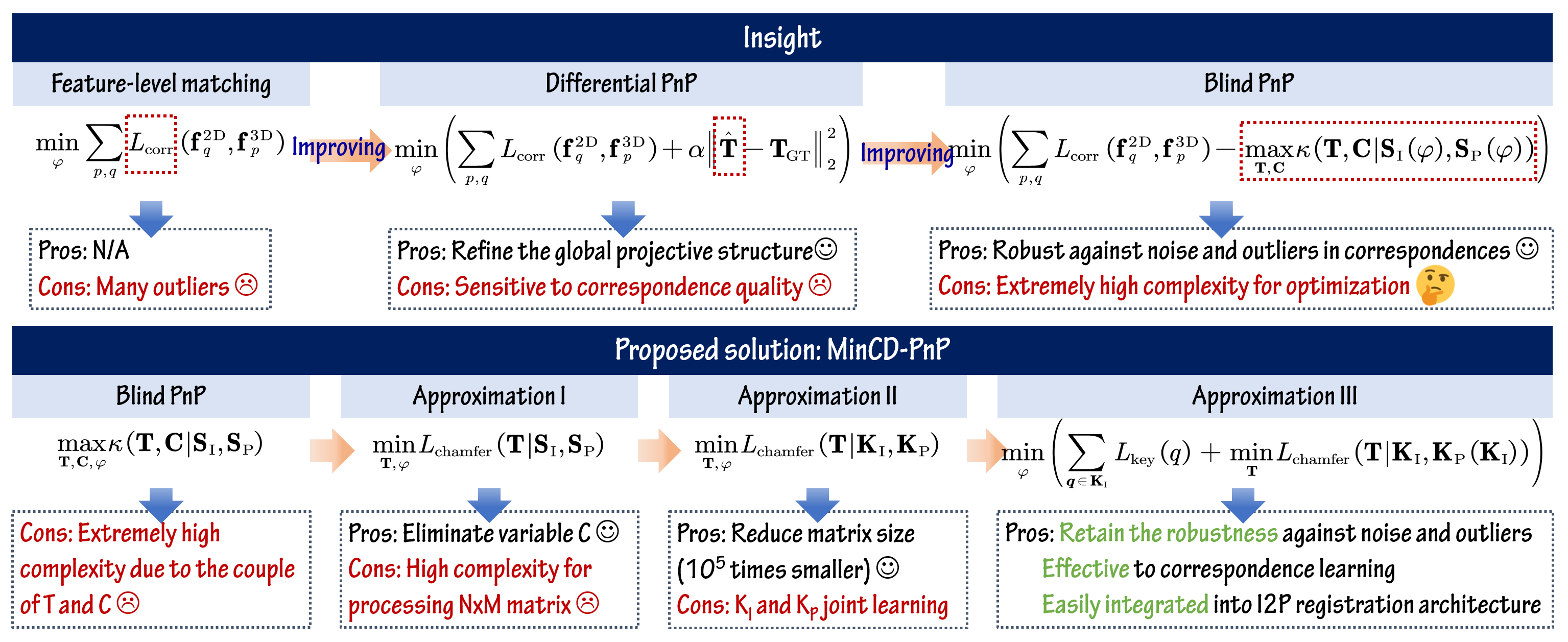}	
		\caption{Motivation of the proposed MinCD-PnP. First, we analyze correspondence learning from the optimization perspective and obverse that blind PnP is robust to the correspondence quality. To mitigate the expensive complexity of blind PnP, we simplify blind PnP as a new task MinCD-PnP using a triple approximation strategy.}
	\label{fig:framework}
\end{figure*}

\section{Related Work}

\noindent \textbf{I2P registration}. Most I2P registration methods rely on deep learning, as DNNs help bridge the modality gap between images and point clouds. They match features in a pixel-to-point manner. In 2019, Feng \textit{et al.} introduced the first deep learning based method for I2P registration, training a DNN to learn 3D keypoints descriptors \cite{2d3d-match}. Li and Lee \cite{deep-i2p} developed DeepI2P, which enhances the feature representation through global feature interaction. Ren \textit{et al.} \cite{corr-tcsvt} further refined this approach in 2023. Building on the image registration method D2-Net \cite{d2-net}, Wang \textit{et al.} \cite{p2-net} developed P2-Net, which jointly learns 2D-3D keypoints and their descriptors. Circle loss \cite{circle-loss} was used to alleviate the extreme imbalance between inliers and outliers. Li \textit{et al.} \cite{2d3d-matr} followed the point cloud registration architecture GeoTrans \cite{geotrans} to develop 2D3D-MATR, which outperformed P2-Net \cite{p2-net}. This work was further improved by Wu \textit{et al.} \cite{diff-reg} in 2024 by integrating a diffusion model \cite{diffusion_model} to iteratively denoise correspondence matrix. In 2024, An \textit{et al.} \cite{an-new-add-1} introduced Proj-ICP, a non-learning algorithm to estimate camera pose by minimizing the 2D-3D contour distances. They also conducted a survey to summarize the I2P registration methods for LiDAR-camera extrinsic calibration \cite{an-survey-1}. Wang \textit{et al.} \cite{freereg} designed an architecture, FreeReg which utilized the pre-trained vision fundamental models to minimize the modality difference between images and point clouds. Based on the above discussions, most current methods \textbf{follows a pixel-to-point feature-matching manner} to establish correspondences.

\noindent \textbf{Learning correspondences with PnP}. Recent research has highlighted the 
absence of the geometrical constraint in I2P registration, leading to the development of differential PnP for improved correspondence learning. In 2023, Zhou \textit{et al.} \cite{diff_reg_match} explored the effect of end-to-end probabilistic PnP
(EPro-PnP) \cite{Epro-PnP} on the 2D-3D correspondence learning task. Although EPro-PnP is robust to correspondence noise, its performance becomes unstable in the presence of excessive outliers. In 2024, Wu \textit{et al.} \cite{diff-reg} regarded correspondence learning as a denoising procedure and combined the diffusion model with differential PnP to refine 2D-3D correspondences. To make differential PnP more robust to correspondence noise and outliers, Campbell \textit{et al.} \cite{blind-pnp} were the first to study blind PnP and designed a weighted differential blind PnP layer based on a declarative network \cite{Declarative_network}. In their work \cite{blind-pnp}, RANSAC-based PnP \cite{p3p} is used to filter correspondences with large noises, and declarative network computes the loss backward gradient of RANSAC-based PnP layer. Although work \cite{blind-pnp} is robust to correspondence noise and outliers, the loss gradient from  filtered correspondences provide limited benefits to the overall I2P architecture. Thus, \textbf{an effective differential PnP for I2P registration} is still an open problem.


\section{Problem Formulation and Analysis}

In this section, we revisit I2P registration from the optimization perspective and analyze the bottleneck of 2D-3D correspondence learning (as illustrated in Fig. \ref{fig:framework}). For a given pixel $q\in\mathcal{I}$ and a point $p\in\mathcal{P}$, their correspondence $\langle q,p \rangle$ is determined using feature-level matching \cite{2d3d-matr,p2-net,freereg}:

\begin{equation}
d(\mathbf{f}_q^{\text{2D}}, \mathbf{f}_p^{\text{3D}}) \leq \delta \Rightarrow \langle q,p \rangle \text{ is a correspondence}
\end{equation}

\begin{equation}
\mathbf{F}_\text{I}, \mathbf{F}_\text{P} = \varphi(\mathcal{I}, \mathcal{P})
\end{equation}

\noindent where $d(\cdot,\cdot)$ represents the per-feature normalized $L_2$ distance, and $\delta$ is a predefined threshold. The features $\mathbf{f}_q^{\text{2D}}$ and $\mathbf{f}_p^{\text{3D}}$ on $q$ and $p$ are extracted from $\mathbf{F}_\text{I}$ and $\mathbf{F}_\text{P}$, respectively, and $\varphi$ denotes the neural network used for I2P registration. It is learned by the following optimization problem:

\begin{equation}
\varphi^{\star} = \arg\min_{\varphi} \sum_{p,q}L_{\text{corr}}(\mathbf{f}_q^{\text{2D}}, \mathbf{f}_p^{\text{3D}})
\end{equation}

\noindent where $p$, $q$ are pixel-to-point pair that satisfies Eq. (1). $L_{\text{corr}}$ is the common correspondence loss, such as circle loss \cite{circle-loss}, because it helps mitigate the severe imbalance between inliers and outliers \cite{2d3d-matr,p2-net}.

The optimization in Eq. (3) is suboptimal because it neglects the projective constraint of $\langle q,p \rangle$. A valid correspondence $\langle q,p \rangle$ must satisfy $q=\pi(\mathbf{T}p)$, where $\pi(\cdot)$ represents the camera projection operator \cite{camera_model}. $\mathbf{T}$ is the transformation from the point cloud to the camera coordinate system. Differential PnP based methods enforce the projective constraints \cite{diff-reg, diff_reg_match}, refining Eq. (3) as:

\begin{equation}
\min_{\varphi} \left(\sum_{p,q}L_{\text{corr}}(\mathbf{f}_q^{\text{2D}}, \mathbf{f}_p^{\text{3D}}) + \alpha \Vert \mathbf{\hat{T}} - \mathbf{T} \Vert_2^2\right)
\end{equation}

\begin{equation}
\mathbf{\hat{T}} = \arg\min_{\mathbf{T}} \sum_{p,q} \mathbb{I}(d(\mathbf{f}_q^{\text{2D}}, \mathbf{f}_p^{\text{3D}}) \leq \delta) \cdot \Vert q-\pi(\mathbf{T}   p) \Vert_2^2
\end{equation}

\noindent where $\Vert q-\pi(\mathbf{T}p) \Vert_2$ computes the re-projection error of correspondence $\langle q,p \rangle$. $\mathbb{I}(x)$ is an indicator function that outputs $1$ if $x$ is true, or $0$ if $x$ is false. Equations (4) and (5) are {coupled} optimization problems, and $\alpha$ is the loss weight. In Eq. (4), the term $\Vert \mathbf{\hat{T}} - \mathbf{T} \Vert_2^2$ enforces global geometrical consistency and improves the accuracy of estimated correspondences. However, solving Eq. (5) is highly sensitive to correspondence noise and outliers  \cite{pnp-sensitive}. Moreover, the network $\varphi$ inevitably predicts outliers and noised inliers, making it a challenge for current differential PnP methods to improve correspondence learning effectively.





\section{Proposed Method}

\subsection{Motivation}

In this paper, we attempt to improve 2D-3D correspondence learning with blind PnP. Its overview is provided in Fig. \ref{fig:framework}. With the blind PnP cost function \cite{go-blind-pnp}, we revise Eq. (3) as:

\begin{equation}
\begin{aligned}
\min_{\varphi} \left(\sum_{p,q}L_{\text{corr}}(\mathbf{f}_q^{\text{2D}}, \mathbf{f}_p^{\text{3D}}) - \max_{\mathbf{T},\mathbf{C}} \kappa(\mathbf{T},\mathbf{C}|\mathbf{S}_\text{I}(\varphi), \mathbf{S}_\text{P}(\varphi))
\right) \\
\text{s.t.}\,\, \mathbf{T}\in \mathbf{SE}(3), \mathbf{C}\in\mathbb{B}^{M\times N}, \mathbf{S}_{\text{I}}=\{q_i\}_{i=1}^M, \mathbf{S}_{\text{P}}=\{p_i\}_{i=1}^N
\end{aligned}
\end{equation}

\begin{equation}
\kappa(\mathbf{T},\mathbf{C}|\mathbf{S}_\text{I}, \mathbf{S}_\text{P}) = \sum_{\langle q,p \rangle\in\mathbf{C}}  \mathbb{I}(\Vert q - \pi(\mathbf{T}  p) \Vert_2^2\leq\tau)
\end{equation}

\noindent where $\mathbf{S}_{\text{I}}(\varphi)$ and $\mathbf{S}_{\text{P}}(\varphi)$ are pixel and point sets of the candidate correspondences, which are sampled from $\mathbf{F}_{\text{I}}$ and $\mathbf{F}_{\text{P}}$ via Eq. (1). As $\mathbf{F}_{\text{I}}$ and $\mathbf{F}_{\text{P}}$ are learned from $\varphi$, $\mathbf{S}_{\text{I}}(\varphi)$ and $\mathbf{S}_{\text{P}}(\varphi)$ can be regarded as the functions of $\varphi$. For the discussion simplicity, $\mathbf{S}_{\text{I}}(\varphi)$ and $\mathbf{S}_{\text{P}}(\varphi)$ are simplified as $\mathbf{S}_{\text{I}}$ and $\mathbf{S}_{\text{P}}$. $\mathbf{C}$ is a boolean $M\times N$ matrix to denote the correspondences between $\mathbf{S}_{\text{I}}$ and $\mathbf{S}_{\text{P}}$. $\kappa(\mathbf{T},\mathbf{C}|\mathbf{S}_\text{I}, \mathbf{S}_\text{P})$ denotes the inlier number, and $\tau$ is a pixel threshold to determine whether the correspondence is an inlier. Blind PnP is robust to correspondence noise and outliers via jointly optimizing $\mathbf{T}$ and $\mathbf{C}$. However,  $\kappa(\mathbf{T},\mathbf{C}|\mathbf{S}_\text{I}, \mathbf{S}_\text{P})$ is an optimization problem with extremely high complexity \cite{yangheng-1}, so that blind PnP cannot be directly used for correspondence learning.  

\subsection{MinCD-PnP formulation}

To address the above issue, we aim to simplify blind PnP as MinCD-PnP via a triple approximation technique.



\subsubsection{Approximation I: from inlier maximization to  Chamfer distance minimization}

First, we aim to approximate the inlier maximization cost function $\kappa(\mathbf{T},\mathbf{C}|\mathbf{S}_\text{I}, \mathbf{S}_\text{P})$ as a lightweight Chamfer distance minimization. To reach this goal, we study an inequality:

\begin{equation}
\begin{aligned}
\max_{\mathbf{T},\mathbf{C}} \kappa(\mathbf{T},\mathbf{C}|\mathbf{S}_\text{I}, \mathbf{S}_\text{P}) &\leq \max_{\mathbf{T}} \kappa(\mathbf{T},\mathbf{C}^\star|\mathbf{S}_\text{I}, \mathbf{S}_\text{P}) \\
&\leq \max_{\mathbf{T}} \kappa^\star(\mathbf{T}^\star |\mathbf{S}_\text{I}, \mathbf{S}_\text{P}) \\
\end{aligned}
\end{equation}

\begin{equation}
\begin{aligned}
\kappa^\star(\mathbf{T} |\mathbf{S}_\text{I}, \mathbf{S}_\text{P}) &= \sum_{q\in\mathbf{S}_\text{I}} \mathbb{I}(\min_{p\in\mathbf{S}_\text{P}} \Vert q - \pi(\mathbf{T}  p) \Vert_2^2 \leq \tau) \\
&+ \sum_{p\in\mathbf{S}_\text{P}} \mathbb{I}(\min_{q\in\mathbf{S}_\text{I}} \Vert q - \pi(\mathbf{T}  p) \Vert_2^2 \leq \tau)
\end{aligned}
\end{equation}

\begin{figure*}[t]
	\centering
		\includegraphics[width=1.0\linewidth]{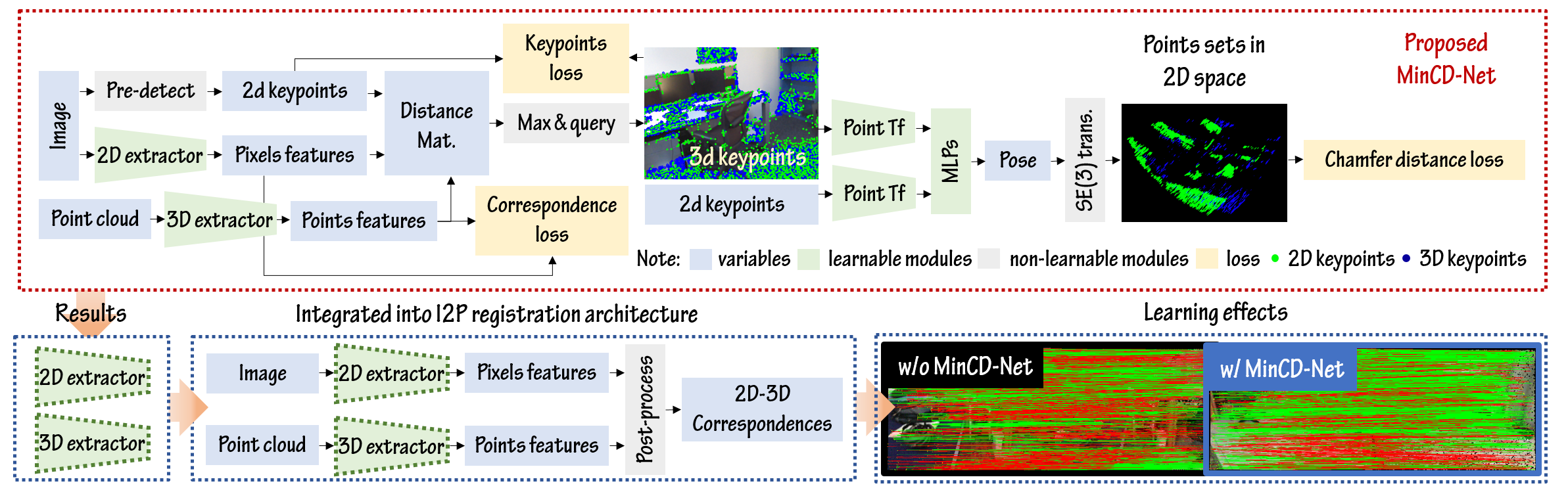}	
		\caption{Proposed 2D-3D correspondence learning module, MinCD-Net. It converts the optimization in Eq. (16) as a \textbf{multi-task learning mechanism}. MinCD-Net can be integrated into existing I2P registration architecture.}
	\label{fig:method}
\end{figure*}

\noindent where $\mathbf{C}^\star$ is the optimal correspondence matrix. It is trivial that $\kappa(\mathbf{T},\mathbf{C}^\star|\mathbf{S}_\text{I}, \mathbf{S}_\text{P})\geq\kappa(\mathbf{T},\mathbf{C}|\mathbf{S}_\text{I}, \mathbf{S}_\text{P})$. We mainly explain the last term in inequality (8).  For a correspondence $\langle q,p \rangle\in\mathbf{C}^\star$, based on the above assumption, we both have $q=\arg\min_{q\in\mathbf{S}_\text{I}} \Vert q - \pi(\mathbf{T}  p) \Vert_2^2$ and $p=\arg\min_{p\in\mathbf{S}_\text{P}} \Vert q - \pi(\mathbf{T}  p) \Vert_2^2$. And $2\kappa(\mathbf{T}^\star,\mathbf{C}^\star|\mathbf{S}_\text{I}, \mathbf{S}_\text{P})=\kappa^\star(\mathbf{T}^\star |\mathbf{S}_\text{I}, \mathbf{S}_\text{P})=2N$, where $\mathbf{T}^\star$ is the optimal  pose. It leads to the last term in inequality (8). Using the inequality (8), we convert the inliers maximization cost function in Eq. (6) as a Chamfer distance minimization cost function:

\begin{equation}
\begin{aligned}
\min_{\varphi} \left(\sum_{p,q}L_{\text{corr}}(\mathbf{f}_q^{\text{2D}}, \mathbf{f}_p^{\text{3D}}) + \min_{\mathbf{T}} L_{\text{Chamfer}}(\mathbf{T}|\mathbf{S}_\text{I}, \mathbf{S}_\text{P})
\right)
\end{aligned}
\end{equation}

\begin{equation}
\begin{aligned}
L_{\text{Chamfer}}(\mathbf{T}|\mathbf{S}_\text{I}, \mathbf{S}_\text{P}) &= \sum_{q\in \mathbf{S}_\text{I}} \min_{p\in\mathbf{S}_\text{P}} \Vert q - \pi(\mathbf{T}  p) \Vert_2^2 \\
&+ \sum_{p\in\mathbf{S}_\text{P}} \min_{q\in\mathbf{S}_\text{I}} \Vert q - \pi(\mathbf{T}  p) \Vert_2^2  \\
\end{aligned}
\end{equation}

The advantage of Eq. (10) is the \textbf{elimination} of  $M\times N$ boolean matrix $\mathbf{C}$, which significantly reduces computation complexity. 

\subsubsection{Approximation II: reducing complexity in Chamfer distance optimization with keypoints}

In the second stage, we introduce the further refinements to improve the optimization efficiency of Eq. (10). Given that an image typically contains $10^6$ pixels and a point cloud typically contains $10^5$ points, $M\times N$ can exceed $10^{11}$, leading to a prohibitively expensive Chamfer distance computation. To address this problem, we sample the representative keypoints $\mathbf{K}_\text{I}=\{q_i\}_{i=1}^{M_0}$ and $\mathbf{K}_\text{P}=\{p_i\}_{i=1}^{N_0}$\footnote{As shown in Eq. (6), $\mathbf{S}_\text{I}$ and $\mathbf{S}_\text{P}$ are functions of $\varphi$, so that $\mathbf{K}_\text{I}$ and $\mathbf{K}_\text{P}$ are also functions of $\varphi$. It means that $\mathbf{K}_\text{I}$ and $\mathbf{K}_\text{P}$ are learned from $\varphi$.} from $\mathbf{S}_\text{I}$ and $\mathbf{S}_\text{P}$, and revise Eq. (10) as:

\begin{equation}
\begin{aligned}
\min_{\varphi} \left(\sum_{p,q}L_{\text{corr}}(\mathbf{f}_q^{\text{2D}}, \mathbf{f}_p^{\text{3D}}) + \min_{\mathbf{T}} L_{\text{Chamfer}}(\mathbf{T}|\mathbf{K}_\text{I}, \mathbf{K}_\text{P})
\right)
\end{aligned}
\end{equation}

The main advantage of Eq. (12) is that the Chamfer distance matrix size is reduced from $M\times N$ to $M_0\times N_0$. As 2D and 3D keypoints number is nearly $10^3$, the matrix size is \textbf{smaller than $10^5$ times}. Although Eq. (12) improves optimization efficiency, a key challenge remains: how to effectively learn the representative $\mathbf{K}_\text{I}$ and $\mathbf{K}_\text{P}$? To ensure that $L_{\text{Chamfer}}(\mathbf{T}|\mathbf{K}_\text{I}, \mathbf{K}_\text{P})$ contributes effectively to $\varphi$, $\mathbf{K}_\text{I}$ and $\mathbf{K}_\text{P}$ should sufficiently represent the 2D and 3D spaces.

\subsubsection{Approximation III: learning 3D keypoints  with the guidance of 2D keypoints}

To deal with the above learning problem of $\mathbf{K}_\text{I}$ and $\mathbf{K}_\text{P}$, we design the third approximation that approximates joint 2D and 3D keypoints learning as a single learning task. We try to learn the 3D keypoints that \textbf{mimic the 2D keypoints distribution}, since jointly learning both with sufficient inliers is a challenging task \cite{2d3d-matr}. In this scheme, $\mathbf{K}_\text{I}$ is pre-detected using a pre-trained model or non-learning algorithm. Existing 2D keypoint detection methods can ensure that $\mathbf{K}_\text{I}$ represents the 2D image. Then, we design a 2D keypoints guided 3D keypoints learning scheme:

\begin{equation}
\begin{aligned}
&\min_{\varphi} \sum_{q\in\mathbf{K}_{\text{I}}} \Vert q-\pi(\mathbf{T}p) \Vert_2 \\
\text{s.t.} \,\, p &= \arg\min_{p\in\mathcal{P}} d(\mathbf{f}_q^{\text{2D}}, \mathbf{f}_p^{\text{3D}}),\,\, q\in \mathbf{K}_\text{I} \\
\end{aligned}
\end{equation}

However, learning with Eq. (13) is ineffective, because some of 2D keypoints without salient features are difficult to find their corresponding 3D keypoints. It makes the loss in Eq. (13) unstable. So, we approximate Eq. (13) as:

\begin{equation}
\begin{aligned}
\min_{\varphi} &\sum_{q\in\mathbf{K}_{\text{I}}} L_{\text{key}}(q) \\
= &\sum_{q\in\mathbf{K}_{\text{I}}} -\mathbb{I}(\Vert q - \pi(\mathbf{T}_{\text{gt}}  p^\star_q) \Vert_2^2 \leq \tau) \cdot \mathbb{I}(s_q^\star \leq s_{\text{th}})
\end{aligned}
\end{equation}

\begin{equation}
\begin{aligned}
p^\star_q &= \arg\min_{p\in \mathcal{P}} \{d(\mathbf{f}_q^{\text{2D}}, \mathbf{f}_1^{\text{3D}}),...d(\mathbf{f}_q^{\text{2D}}, \mathbf{f}_{p}^{\text{3D}})...,d(\mathbf{f}_q^{\text{2D}}, \mathbf{f}_{N_0}^{\text{3D}})\} \\
s^\star_q &= \min_{p\in \mathcal{P}} \{d(\mathbf{f}_q^{\text{2D}}, \mathbf{f}_1^{\text{3D}}),...d(\mathbf{f}_q^{\text{2D}}, \mathbf{f}_{p}^{\text{3D}})...,d(\mathbf{f}_q^{\text{2D}}, \mathbf{f}_{N_0}^{\text{3D}})\} \\
\end{aligned}
\end{equation}

\noindent where the term $\min \{d(\mathbf{f}_q^{\text{2D}}, \mathbf{f}_1^{\text{3D}}),...,d(\mathbf{f}_q^{\text{2D}}, \mathbf{f}_{N_0}^{\text{3D}})\}$ approximates $d(\mathbf{f}_q^{\text{2D}}, \mathbf{f}_p^{\text{3D}})$. This implies that Eq. (15) aims to learn a set of 3D keypoints that best \textbf{approximate} the detected 2D keypoints in an error bound of $\tau$. $s_{\text{th}}$ is a threshold used to filter out low-confidence 3D keypoints.

Following the triple  approximation, we formulate the proposed scheme as a new optimization problem that minimizes Chamfer distance and 3D keypoints learning losses:

\begin{equation}
\begin{aligned}
\varphi^{\star} &= \arg\min_{\varphi} \left(\sum_{p,q}L_{\text{corr}}(\mathbf{f}_q^{\text{2D}}, \mathbf{f}_p^{\text{3D}})\right. + \lambda_1 \sum_{q\in\mathbf{K}_{\text{I}}} L_{\text{key}}(q) \\
&\left. +\lambda_2 \min_{\mathbf{T}} L_{\text{Chamfer}}(\mathbf{T}|\mathbf{K}_\text{I}, \mathbf{K}_\text{P}(\mathbf{K}_\text{I})) \right)
\end{aligned}
\end{equation}

\noindent where $\lambda_1$ and $\lambda_2$ are the loss weights. $\mathbf{K}_\text{P}(\mathbf{K}_\text{I})$ denotes that 3D keypoints are learned from 2D keypoints via Eq. (15). 




\subsection{Correspondence learning with MinCD-PnP}

In Sec. 4.2, we have modeled the correspondence learning as a MinCD-PnP problem. To effectively address MinCD-PnP, we propose a lightweight multi-task learning module, MinCD-Net, as shown in Fig. \ref{fig:method}. Its core is to predict 3D keypoints and compute multi-task losses in Eq. (16). 



\begin{table*}[!ht]
\scriptsize
\centering
\caption{I2P registration performance for cross-scene generalization on the 7-Scenes datasets. Here $\dag$ represents the average metrics across the unseen scenes. MinCD-Net achieves higher IR and RR than other methods in most of scenes. Bold indicates the best performance.}
\begin{tabular}{c|c|cccccc|c}
\hline
\textbf{IR} & Chess$\rightarrow$Chess & Chess$\rightarrow$Fire & Chess$\rightarrow$Heads & Chess$\rightarrow$Office & Chess$\rightarrow$Pumpkin & Chess$\rightarrow$Kitchen & Chess$\rightarrow$Stairs & AVG$^\dag$ \\
\hline
P2-Net    & 0.516 & 0.436 & 0.330 & 0.414 & {{0.421}} & 0.405 & 0.251 & 0.376 \\
MATR & 0.761 & 0.455 & 0.359 & 0.420 & 0.411 & 0.390 & 0.288 & 0.387 \\
+Diff. PnP & 0.753 & 0.462 & 0.364 & 0.427 & 0.424 & 0.402 & 0.285 & 0.394 \\
+BPnPNet & 0.747 & 0.492 & 0.397 & 0.476 & \textcolor{blue}{\textbf{0.450}} & 0.365 & 0.342 & 0.420 \\
 +MinCD-Net   & \textcolor{blue}{\textbf{0.816}} & \textcolor{blue}{\textbf{0.542}} & \textcolor{blue}{\textbf{0.424}} & \textcolor{blue}{\textbf{0.502}} & {{0.408}} & \textcolor{blue}{\textbf{0.416}} & \textcolor{blue}{\textbf{0.379}} & \textcolor{blue}{\textbf{0.445}} \\
\hline
\textbf{RR} & Chess$\rightarrow$Chess & Chess$\rightarrow$Fire & Chess$\rightarrow$Heads & Chess$\rightarrow$Office & Chess$\rightarrow$Pumpkin & Chess$\rightarrow$Kitchen & Chess$\rightarrow$Stairs & AVG$^\dag$ \\
\hline
P2-Net    & 0.875 & 0.536 & 0.162 & 0.672 & 0.561 & 0.563 & 0.293 & 0.464 \\
MATR &  \textcolor{blue}{\textbf{1.000}} & 0.537 & 0.167 & 0.759 & {{0.581}} & 0.612 & 0.214 & 0.478 \\
+Diff. PnP &  \textcolor{blue}{\textbf{1.000}} & 0.556 & 0.184 & 0.767 & {{0.585}} & 0.622 & 0.226 & 0.490 \\
+BPnPNet & \textcolor{blue}{\textbf{1.000}} & 0.665 & 0.224 & 0.778 & \textcolor{blue}{\textbf{0.660}} & 0.601 & 0.142 & 0.512 \\
 +MinCD-Net   & {{0.985}} & \textcolor{blue}{\textbf{0.671}} & \textcolor{blue}{\textbf{0.250}} & \textcolor{blue}{\textbf{0.869}} & {{0.574}} & \textcolor{blue}{\textbf{0.619}} & \textcolor{blue}{\textbf{0.571}} & \textcolor{blue}{\textbf{0.592}} \\
\hline
\textbf{IR} & Office$\rightarrow$Office & Office$\rightarrow$Chess & Office$\rightarrow$Fire & Office$\rightarrow$Heads & Office$\rightarrow$Pumpkin & Office$\rightarrow$Kitchen & Office$\rightarrow$Stairs & AVG$^\dag$ \\
\hline
P2-Net    & 0.506 & 0.416 & 0.413 & 0.403 & 0.434 & 0.386 & 0.308 & 0.393 \\
MATR & 0.645 & 0.498 & 0.491 & 0.521 & 0.442 & 0.448 & 0.338 & 0.456 \\
+Diff. PnP & 0.653 & 0.502 & 0.497 & 0.532 & 0.439 & 0.457 & 0.351 & 0.463 \\
+BPnPNet & 0.666 & 0.554 & 0.565 & 0.473 & 0.472 & 0.454 & 0.389 & 0.486 \\
 +MinCD-Net   & \textcolor{blue}{\textbf{0.783}} & \textcolor{blue}{\textbf{0.660}} & \textcolor{blue}{\textbf{0.642}} & \textcolor{blue}{\textbf{0.536}} & \textcolor{blue}{\textbf{0.550}} & \textcolor{blue}{\textbf{0.546}} & \textcolor{blue}{\textbf{0.471}} & \textcolor{blue}{\textbf{0.568}} \\
\hline
\textbf{RR} & Office$\rightarrow$Office & Office$\rightarrow$Chess & Office$\rightarrow$Fire & Office$\rightarrow$Heads & Office$\rightarrow$Pumpkin & Office$\rightarrow$Kitchen & Office$\rightarrow$Stairs & AVG$^\dag$ \\
\hline
P2-Net    & 0.769 & 0.566 & 0.661 & 0.232 & 0.577 & 0.532 & 0.234 & 0.510 \\
MATR & {{0.940}} & 0.660 & 0.556 & {{0.417}} & 0.395 & 0.636 & 0.286 & 0.491 \\
+Diff. PnP & {{0.947}} & 0.672 & 0.559 & \textcolor{blue}{\textbf{0.422}} & 0.402 & 0.648 & 0.301 & 0.501 \\
+BPnPNet & 0.848 & 0.708 & \textcolor{blue}{\textbf{0.781}} & 0.144 & 0.660 & 0.750 & 0.429 & 0.578 \\
 +MinCD-Net   & \textcolor{blue}{\textbf{0.980}} & \textcolor{blue}{\textbf{0.769}} & {{0.726}} & {{0.250}} & \textcolor{blue}{\textbf{0.681}} & \textcolor{blue}{\textbf{0.810}} & \textcolor{blue}{\textbf{0.643}} & \textcolor{blue}{\textbf{0.647}} \\
\hline
\textbf{IR} & Kitchen$\rightarrow$Kitchen & Kitchen$\rightarrow$Chess & Kitchen$\rightarrow$Fire  & Kitchen$\rightarrow$Office & Kitchen$\rightarrow$Heads & Kitchen$\rightarrow$Pumpkin & Kitchen$\rightarrow$Stairs & AVG$^\dag$ \\
\hline
P2-Net    & 0.678 & 0.516 & 0.512 & 0.504 & 0.506 & 0.555 & 0.358 & 0.491 \\
MATR & {{0.717}} & 0.571 & 0.594 & 0.537 & 0.538 & 0.612 & 0.370 & 0.537 \\
+Diff. PnP & {{0.723}} & 0.576 & 0.602 & 0.545 & 0.546 & 0.627 & 0.382 & 0.546 \\
+BPnPNet & 0.693 & 0.562 & 0.557 & 0.530 & 0.562 & 0.576 & 0.409 & 0.532 \\
 +MinCD-Net   & \textcolor{blue}{\textbf{0.778}} & \textcolor{blue}{\textbf{0.617}} & \textcolor{blue}{\textbf{0.598}} & \textcolor{blue}{\textbf{0.540}} & \textcolor{blue}{\textbf{0.573}} & \textcolor{blue}{\textbf{0.636}} & \textcolor{blue}{\textbf{0.445}} & \textcolor{blue}{\textbf{0.568}} \\
\hline
\textbf{RR} & Kitchen$\rightarrow$Kitchen & Kitchen$\rightarrow$Chess & Kitchen$\rightarrow$Fire  & Kitchen$\rightarrow$Office & Kitchen$\rightarrow$Heads & Kitchen$\rightarrow$Pumpkin & Kitchen$\rightarrow$Stairs & AVG$^\dag$ \\
\hline
P2-Net    & 0.851 & 0.857 & 0.583 & 0.250 & 0.769 & 0.611 & 0.429 & 0.621 \\
MATR & 0.901 & 0.872 & 0.778 & 0.667 & 0.723 & 0.698 & 0.500 & 0.706 \\
+Diff. PnP & 0.918 & 0.885 & 0.783 & 0.685 & 0.741 & 0.703 & 0.532 & 0.722 \\
+BPnPNet & \textcolor{blue}{\textbf{0.923}} & \textcolor{blue}{\textbf{0.954}} & 0.849 & 0.650 & 0.717 & 0.830 & 0.714 & 0.785 \\
 +MinCD-Net   & {{0.875}} & {{0.846}} & \textcolor{blue}{\textbf{0.904}} & \textcolor{blue}{\textbf{0.683}} & \textcolor{blue}{\textbf{0.798}} & \textcolor{blue}{\textbf{0.872}} & \textcolor{blue}{\textbf{0.786}} & \textcolor{blue}{\textbf{0.814}} \\
\hline
\end{tabular}
\label{tab:res_1}
\end{table*}

\subsubsection{General architecture of I2P registration}

Before introducing the proposed MinCD-Net, we briefly describe the architecture of the I2P registration network for better clarity. As shown in the left part of Fig. \ref{fig:method}, the current network incorporates two feature extractors for learning images and point clouds features $\mathbf{F}_{\text{I}}$ and $\mathbf{F}_{\text{P}}$. In the previous work \cite{2d3d-matr}, image extractor is ResNet \cite{resnet} and point cloud extractor is KPConv \cite{kpconv}. A key step in I2P registration is post-processing. Li \textit{et al.} \cite{2d3d-matr} designed a two-stage matching scheme inspired by GeoTrans \cite{geotrans}. In the first stage, 2D and 3D patch features (i.e., obtained from extractors) are used for the 2D-3D patches matching. Then, for every matched patch pair, correspondences are determined using Eq. (1). In all, $\varphi$ in Eq. (2) represents two feature extractors, and the detail of $L_{\text{corr}}$ can refer to literature \cite{2d3d-matr,p2-net}.

\subsubsection{Keypoints loss and Chamfer loss computation}

We provide the computational detail of $L_{\text{key}}(q)$ in Eqs. (13-15). By computing the L2 distance between each 2D keypoint to each 3D point feature, we obtain a $M_0\times N$ distance matrix $\mathbf{D}=(d_{ij})$ with $d_{ij}=d(\mathbf{f}_i^{\text{2D}}, \mathbf{f}_j^{\text{3D}})$. Using the Pytorch API function \texttt{min}, elements in Eq. (15) are obtained. To efficiently compute $\mathbb{I}(\Vert q - \pi(\mathbf{T}_{\text{gt}}   p^\star_q) \Vert_2^2 \leq \tau)$, we pre-compute the overlapping mask in $\mathcal{P}$, converting $L_{\text{key}}(q)$ as a loss function based on the intersection of union (IoU) of two sets. We empirically set $s_{\text{th}}$ to $e^{-0.4}$ for the best performance.

Next, we analyze the Chamfer loss $L_{\text{Chamfer}}(\mathbf{T}|\mathbf{K}_\text{I}, \mathbf{K}_\text{P})$ in Eq. (16). Minimizing $L_{\text{Chamfer}}(\mathbf{T}|\mathbf{K}_\text{I}, \mathbf{K}_\text{P})$ during training is computationally expensive. We predict $\mathbf{T}$ from $\mathbf{K}_\text{I}$ and $\mathbf{K}_\text{P}$ in an end-to-end manner where $L_{\text{Chamfer}}(\mathbf{T}|\mathbf{K}_\text{I}, \mathbf{K}_\text{P})$ serves as a loss function. In MinCD-Net, we utilize point transformer (PointTf) \cite{pointTf} to encode 2D and 3D keypoint features\footnote{2D features contain pixels' 2D bearing vectors and features obtained from 2D extractor. 3D features contain points' 3D coordinates and features obtained from 3D extractors.}, and then compute the global 2D and 3D features. By concatenating these global features, we use a series of multilayer perceptrons (MLPs) to estimate $\mathbf{T}$ \cite{calibnet}. With $\mathbf{T}$, we can transform the coordinates of $\mathbf{K}_\text{P}$ and then compute the Chamfer loss $L_{\text{Chamfer}}(\mathbf{T}|\mathbf{K}_\text{I}, \mathbf{K}_\text{P})$.



\subsubsection{Summary}

We summarize the {effect} of MinCD-Net on I2P registration. First, MinCD-Net is \textbf{robust to the noise and outliers} in the predicted correspondences, because the proposed loss functions (i.e., $L_{\text{key}}(q)$ and $L_{\text{Chamfer}}(\mathbf{T}|\mathbf{K}_\text{I}, \mathbf{K}_\text{P})$) are only related to $\mathbf{K}_\text{I}$ and $\mathbf{K}_\text{P}$. It addresses the limitations of existing differential PnP schemes \cite{blind-pnp, diff-reg, diff_reg_match}. Second, MinCD-Net is \textbf{effective to learning $\varphi$}. Since the pre-detected 2D keypoints can represent the 2D image, $L_{\text{key}}(q)$ ensures that the learned 3D keypoints are close to the pre-detected 2D keypoints. It enforces the loss gradient $\nabla_{\varphi} L_{\text{Chamfer}}(\mathbf{T}|\mathbf{K}_\text{I}, \mathbf{K}_\text{P})$ close related to the pixels and points which represents the whole scene. Thus, the backpropagation of $ L_{\text{Chamfer}}(\mathbf{T}|\mathbf{K}_\text{I}, \mathbf{K}_\text{P})$ contributes more effectively to $\varphi$ compared to existing differential PnP schemes \cite{blind-pnp, diff-reg, diff_reg_match}. Third, MinCD-Net can be \textbf{easily integrated} with existing I2P registration networks, as its inputs are independent of the outputs of I2P registration networks.

\begin{figure*}[ht]
	\centering
		\includegraphics[width=1.0\linewidth]{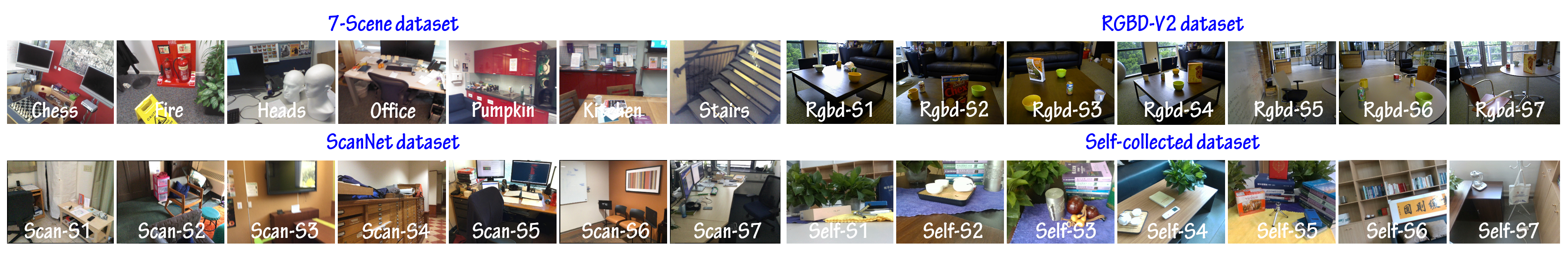}	
		\caption{Example scenes from the 7-Scenes, RGBD-V2, ScanNet, and self-collected datasets (referred to as $\textit{Rgbd}$, $\textit{Scan}$, and $\textit{Self}$).}
	\label{fig:datasets}
\end{figure*}

\begin{table*}[ht]
    \scriptsize
    \centering
        \caption{I2P registration performance for cross-dataset generalization on the multiple datasets, including RGBD-V2, ScanNet, and self-collected datasets. The proposed MinCD-Net outperforms other methods in most of the scenes.}
    \begin{tabular}{c|ccccccc|c}
    \hline
    \textbf{IR} & Kit$\rightarrow$Rgbd-S1 & Kit$\rightarrow$Rgbd-S2 & Kit$\rightarrow$Rgbd-S3 & Kit$\rightarrow$Rgbd-S4 & Kit$\rightarrow$Rgbd-S5 & Kit$\rightarrow$Rgbd-S6 & Kit$\rightarrow$Rgbd-S7 & Average \\
    \hline
    MATR   & 0.351 & 0.353 & 0.336 & 0.316 & 0.250 & 0.209 & 0.222 & 0.291 \\
    +Diff. PnP   & 0.372 & 0.358 & 0.352 & 0.332 & 0.262 & 0.214 & 0.235 & 0.303 \\
    +BPnPNet & 0.396 & 0.378 & 0.375 & 0.377 & 0.230 & 0.194 & 0.258 & 0.315 \\
     +MinCD-Net   & \textcolor{blue}{\textbf{0.427}} & \textcolor{blue}{\textbf{0.415}} & \textcolor{blue}{\textbf{0.405}} & \textcolor{blue}{\textbf{0.412}} & \textcolor{blue}{\textbf{0.310}} & \textcolor{blue}{\textbf{0.296}} & \textcolor{blue}{\textbf{0.329}} & \textcolor{blue}{\textbf{0.371}} \\
    \hline
    \textbf{RR@0.1} & Kit$\rightarrow$Rgbd-S1 & Kit$\rightarrow$Rgbd-S2 & Kit$\rightarrow$Rgbd-S3 & Kit$\rightarrow$Rgbd-S4 & Kit$\rightarrow$Rgbd-S5 & Kit$\rightarrow$Rgbd-S6 & Kit$\rightarrow$Rgbd-S7 & Average \\
    \hline
    MATR & 0.970 & {{0.880}} & 0.871 & 0.741 & 0.480 & 0.449 & 0.458 & 0.692 \\
    +Diff. PnP & 0.972 & {{0.943}} & 0.892 & 0.750 & 0.485 & 0.453 & 0.464 & 0.708 \\
    +BPnPNet & 0.965 & 0.974 & 0.954 & 0.942 & 0.610 & 0.507 & 0.646 & 0.799 \\
     +MinCD-Net   & \textcolor{blue}{\textbf{0.974}} & \textcolor{blue}{\textbf{0.985}} & \textcolor{blue}{\textbf{0.968}} & \textcolor{blue}{\textbf{0.963}} & \textcolor{blue}{\textbf{0.707}} & \textcolor{blue}{\textbf{0.725}} & \textcolor{blue}{\textbf{0.711}} & \textcolor{blue}{\textbf{0.870}} \\
    \hline
    \textbf{IR} & Kit$\rightarrow$Scan-S1 & Kit$\rightarrow$Scan-S2 & Kit$\rightarrow$Scan-S3 & Kit$\rightarrow$Scan-S4 & Kit$\rightarrow$Scan-S5 & Kit$\rightarrow$Scan-S6 & Kit$\rightarrow$Scan-S7 & Average \\
    \hline
    MATR & 0.495 & {{0.550}} & 0.424 & 0.337 & 0.507 & 0.434 & 0.414 & 0.451 \\
    +Diff. PnP & 0.491 & \textcolor{blue}{\textbf{0.552}} & 0.417 & 0.339 & 0.495 & 0.424 & 0.408 & 0.442 \\
    +BPnPNet & 0.504 & 0.511 & 0.426 & 0.324 & 0.529 & 0.427 & 0.405 & 0.446 \\
     +MinCD-Net   & \textcolor{blue}{\textbf{0.517}} & {{0.527}} & \textcolor{blue}{\textbf{0.460}} & \textcolor{blue}{\textbf{0.343}} & \textcolor{blue}{\textbf{0.548}} & \textcolor{blue}{\textbf{0.456}} & \textcolor{blue}{\textbf{0.428}} & \textcolor{blue}{\textbf{0.468}} \\
    \hline
    \textbf{RR@0.05} & Kit$\rightarrow$Scan-S1 & Kit$\rightarrow$Scan-S2 & Kit$\rightarrow$Scan-S3 & Kit$\rightarrow$Scan-S4 & Kit$\rightarrow$Scan-S5 & Kit$\rightarrow$Scan-S6 & Kit$\rightarrow$Scan-S7 & Average \\
    \hline
    MATR & {{0.956}} & {{0.954}} & \textcolor{blue}{\textbf{0.974}} & 0.433 & 0.923 & 0.909 & 0.750 & 0.842 \\
    +Diff. PnP & {{0.932}} & {{0.927}} & {{0.945}} & 0.431 & 0.947 & 0.912 & 0.757 & 0.836 \\
    +BPnPNet & 0.929 & 0.943 & 0.917 & 0.455 & 0.960 & \textcolor{blue}{\textbf{0.923}} & 0.782 & 0.844 \\
     +MinCD-Net   & \textcolor{blue}{\textbf{0.987}} & \textcolor{blue}{\textbf{0.979}} & {{0.905}} & \textcolor{blue}{\textbf{0.720}} & \textcolor{blue}{\textbf{0.962}} & {{0.915}} & \textcolor{blue}{\textbf{0.821}} & \textcolor{blue}{\textbf{0.898}} \\
    \hline
    \textbf{IR} & Kit$\rightarrow$Self-S1 & Kit$\rightarrow$Self-S2 & Kit$\rightarrow$Self-S3 & Kit$\rightarrow$Self-S4 & Kit$\rightarrow$Self-S5 & Kit$\rightarrow$Self-S6 & Kit$\rightarrow$Self-S7 & Average \\
    \hline
    MATR & \textcolor{blue}{\textbf{0.497}} & 0.462 & 0.426 & \textcolor{blue}{\textbf{0.618}} & {{0.507}} & \textcolor{blue}{\textbf{0.619}} & 0.412 & 0.506 \\
    +Diff. PnP & {{0.473}} & 0.453 & 0.421 & {{0.592}} & {{0.516}} & {0.608} & 0.438 & 0.498 \\
    +BPnPNet & 0.462 & 0.442 & 0.415 & 0.572 & 0.513 & 0.598 & 0.495 & 0.499 \\
     +MinCD-Net   & {{0.485}} & \textcolor{blue}{\textbf{0.470}} & \textcolor{blue}{\textbf{0.437}} & {{0.581}} & \textcolor{blue}{\textbf{0.522}} & {{0.604}} & \textcolor{blue}{\textbf{0.514}} & \textcolor{blue}{\textbf{0.516}} \\
    \hline
    \textbf{RR@0.05} & Kit$\rightarrow$Self-S1 & Kit$\rightarrow$Self-S2 & Kit$\rightarrow$Self-S3 & Kit$\rightarrow$Self-S4 & Kit$\rightarrow$Self-S5 & Kit$\rightarrow$Self-S6 & Kit$\rightarrow$Self-S7 & Average \\
    \hline
    MATR & {{0.556}} & 0.389 & 0.333 & {{0.976}} & 0.532 & {{0.964}} & 0.278 & 0.575 \\
    +Diff. PnP & \textcolor{blue}{\textbf{0.564}} & 0.372 & 0.345 & {{0.979}} & 0.545 & \textcolor{blue}{\textbf{0.966}} & 0.306 & 0.582 \\
    +BPnPNet & 0.502 & 0.362 & 0.352 & 0.981 & 0.584 & 0.948 & 0.334 & 0.580 \\
     +MinCD-Net   & {{0.512}} & \textcolor{blue}{\textbf{0.405}} & \textcolor{blue}{\textbf{0.389}} & \textcolor{blue}{\textbf{0.984}} & \textcolor{blue}{\textbf{0.611}} & {{0.952}} & \textcolor{blue}{\textbf{0.389}} & \textcolor{blue}{\textbf{0.606}} \\
    \hline
    \end{tabular}
    \label{tab:res_2}
\end{table*}

\section{Experiments and Discussions}

\subsection{Configurations}

To evaluate the performance of the proposed I2P registration method, we conduct experiments on multi-dataset, including RGBD-V2 \cite{RGBD-dataset}, 7-Scenes \cite{scene-7-dataset}, ScanNet \cite{scan-net}, and the self-collected dataset captured by Intel RealSense depth camera. Examples of scenes are provided in Fig. \ref{fig:datasets}. The train-test data split for RGBD-V2 and 7-Scenes follows previous work \cite{2d3d-matr}, while ScanNet and self-collected datasets are totally utilized for testing. IR and RR are the primary metrics used to evaluate I2P registration. Details of these metrics are given in the appendices of the work \cite{2d3d-matr}. Threshold of IR is 0.05m. RR@X represents the RR threshold at X meters, with a default of 0.05m.


\begin{table*}[ht]
\scriptsize
\centering
\caption{Comparison results of current methods on the RGBD-v2 dataset, evaluated with an RMSE threshold of $0.1$m. $\dag$ denotes that the proposed method has been pre-trained on several indoor datasets, including 7-Scene and ScanNet.}
\begin{tabular}{c|ccccccccccc}
\hline
\textbf{Methods} & MATR & MATR+SN & MATR+D & MATR+Dino & Predator & FreeReg+Kabsch & FreeReg+PnP & Diff-Reg & MinCD-Net & MinCD-Net$^\dag$ \\
\hline
\textbf{IR}       & 0.324 & 0.451 & 0.406 & 0.434 & 0.157 & 0.309 & 0.309 & 0.377 & 0.472 & \textcolor{blue}{\textbf{0.581}} \\
\textbf{RR@0.1}   & 0.564 & 0.770 & 0.668 & 0.744 & 0.302 & 0.341 & 0.573 & 0.862 & 0.823 & \textcolor{blue}{\textbf{0.914}} \\
\hline
\end{tabular}
\label{tab:res_4}
\end{table*}

The implementation of MinCD-Net is discussed. Its inputs include an RGB image with surface normals and RGB point cloud with surface normals. Image surface normals are predicted using the pre-trained model DSINE \cite{dsine}. The extractors in Fig. \ref{fig:method} are ResNet \cite{resnet} and KPConv \cite{kpconv}, where the extractor networks are similar to those in MATR \cite{2d3d-matr}. The threshold $s_{\text{th}}$ in Eq. (14) is set to $e^{-0.4}$. Point transformer in Fig. \ref{fig:method} is the single layer of work \cite{pointTf}. Its key, query, and value inputs are the $128$ dimensional features which are transformed from pixels and points features. To estimate the camera pose, MLPs with two layers, $[256,128]$ and $[128,6]$, are used to predict a $6\times 1$ vector representing the $\mathbf{se}(3)$ of $\mathbf{T}$, and $\mathbf{T}$ is computed via the mapping from $\mathbf{se}(3)$ to $\mathbf{SE}(3)$. We utilize Shi-Tomasi keypoint detection provided by OpenCV API \texttt{Good Features to Track} to extract $\mathbf{K}_\text{I}$ that  are uniformly distributed in the image. MinCD-Net is trained on a single Nvidia RTX 3080 GPU for 40 epochs. In the first 20 epochs, $\lambda_1$ and $\lambda_2$ are set to zero. According to the camera model \cite{camera_model}, the criterion of $\tau$ is:

\begin{equation}
\tau \leq \left( \frac{\text{Threshold of RR}\cdot \max(f_u,f_v)}{d_{\text{max}}} \right)^2
\end{equation}

\noindent where $f_u$ and $f_v$ are camera focal lengths, $d_{\text{max}}$ is the maximum depth. On 7-Scenes dataset \cite{RGBD-dataset}, $f_u=f_v=585.0$ and $d_{\text{max}}=10.0m$. If the RR threshold is 0.05m, $\tau$ is best set to $5$. Besides, $\lambda_1$ and $\lambda_2$ are empirically set to $0.2$ and $0.0001$ for the best performance.

\begin{figure*}[t]
	\centering
		\includegraphics[width=1.0\linewidth]{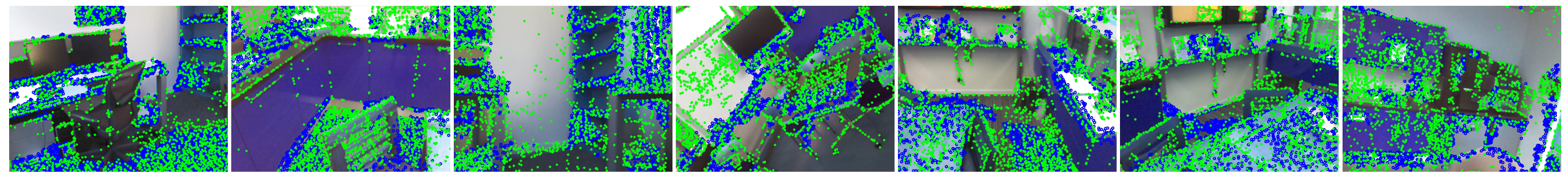}	
		\caption{Visualization of pre-detected 2D keypoints (green dots) and learned 3D keypoints (blue dots). With the proposed sub-optimal learning scheme in Sec. 3.2.3, the learned 3D keypoints exhibit a large overlap with the 2D keypoints.}
	\label{fig:vis-2}
\end{figure*}

\begin{figure*}[!ht]
	\centering
		\includegraphics[width=0.9\linewidth]{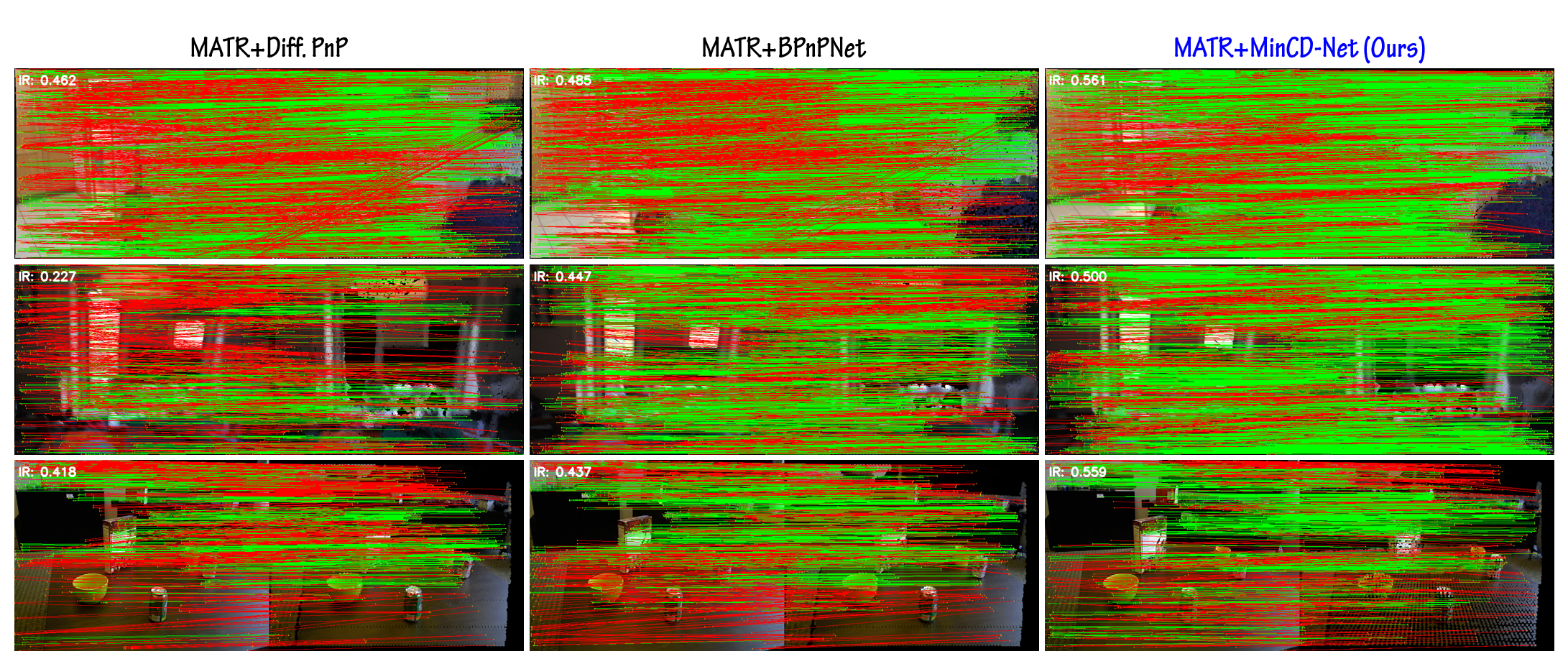}	
		\caption{Visualization of different methods. MinCD-Net achieves the higher correspondence accuracy than other methods.}
	\label{fig:vis-1}
\end{figure*}

\subsection{Methods Comparisons}

To investigate the overall performance of MinCD-Net, we conduct experiments in three different evaluation settings.


\noindent \textbf{Cross-scene generalization}. First, we conduct the cross-scene experiment on the 7-scenes dataset \cite{scene-7-dataset} that contains seven independent indoor scenes. The notation $A\rightarrow B$ indicates that the model is trained on scene $A$ and tested on scene $B$. As the proposed framework falls into the category of differential PnP methods, we mainly compare with two representative methods: Diff. PnP \cite{Epro-PnP} and BPnPNet \cite{blind-pnp}. BPnPNet \cite{blind-pnp} is a previous work that used Blind PnP in correspondence learning. For a fair evaluation, all methods are based on the same baseline, MATR\footnote{MATR\cite{2d3d-matr} is a representative baseline for I2P registration task.} \cite{2d3d-matr}. Thus, we refer to them as MATR+MinCD-Net (ours), MATR+Diff. PnP, and MATR+BPnPNet, respectively. Another classic method, P2-Net \cite{p2-net} is also used for comparison. The results are shown in Table \ref{tab:res_1}. MATR+MinCD-Net has a significant improvement on both the IR and RR metrics than other methods if the training scene is Office. When the training scene is the Chess or Kitchen, MATR+MinCD-Net also outperforms other methods, although the improvement in the IR metric is not significant. From Table \ref{tab:res_1}, MATR+MinCD-Net demonstrates the more robust performance than other methods in the case of Chess$\rightarrow$Stairs, Kitchen$\rightarrow$Stairs, and Kitchen$\rightarrow$Office. More visualization results are provided in Fig. \ref{fig:vis-1}. So, the proposed MinCD-Net achieves both robust and accurate  performance compared to existing differential PnP based methods in the cross-scene setting. 

\noindent \textbf{Cross-dataset generalization}. Next, we evaluate the differential PnP based methods in the cross-dataset setting. The results are shown in  Table \ref{tab:res_2}. On the RGBD-V2 dataset \cite{RGBD-dataset}, the IR metric of MinCD-Net outperforms other methods. On the ScanNet dataset \cite{scan-net}, all methods exhibit the similar performance in the IR metric, but MATR+MinCD-Net learns high-quality correspondences (as seen in the RR metric for Office$\rightarrow$Scan-s4). The self-collected dataset is the most challenging dataset, leading to the poor RR metrics for all methods. Even in these challenging scenes, our method achieves the highest average IR and RR, indicating its effectiveness in the cross-dataset setting. 

\begin{table}[t]
\scriptsize
\centering
\caption{Additional comparison results on the ScanNet dataset. $\dag$ indicates that model was trained on the Kitchen scene with an RR threshold of $0.05$m, \textbf{stricter} than $0.3$m.}
\begin{tabular}{c|cccccc}
\hline
 & P2-Net$^\dag$ & MATR$^\dag$ & LCD & Glue & FreeReg & MinCD-Net$^\dag$ \\
\hline
\textbf{IR}       & 0.303 & 0.451 & 0.307 & 0.184 & \textcolor{blue}{\textbf{0.568}} & 0.468 \\
\textbf{RR}   & 0.711 & 0.842 & N/A & 0.065 & 0.780 & \textcolor{blue}{\textbf{0.898}} \\
\hline
\end{tabular}
\label{tab:res_3}
\end{table}

\noindent \textbf{Standard comparison}. After that, we evaluate the state-of-the-art methods, including 2D3D-MATR \cite{2d3d-matr}, Predator \cite{Predator}, FreeReg \cite{freereg}, and Diff-Reg \cite{diff-reg} on the RGBD-V2 dataset \cite{RGBD-dataset}. All models are trained and tested on the same data split of the RGBD-V2 dataset \cite{RGBD-dataset}. Results are provided in Table \ref{tab:res_4}. +SN/+D/+Dino denotes the use of surface normals \cite{dsine}, monocular depth \cite{depth_anything}, and the pre-trained Dino v2 backbone \cite{DINOv2}. +Kabsch/+PnP denotes the use of Kabsch \cite{Kabsch} and EPnP \cite{epnp} algorithms in outliers removal. Diff-Reg \cite{diff-reg} exploits the EPro-PnP \cite{Epro-PnP} in the correspondence learning. MATR+MinCD-Net outperforms existing methods. Moreover, we conduct an extra comparison on the ScanNet dataset \cite{scan-net} with other approaches, such as LCD \cite{lcd}, Superglue (Glue) \cite{superglue}, and FreeReg \cite{freereg}. Results are shown in Table \ref{tab:res_3}. Even with a strict RR threshold, MinCD-Net still achieves a higher RR than FreeReg \cite{freereg}.

\noindent \textbf{Results analysis}. We analyze why MinCD-Net outperforms Diff. PnP \cite{Epro-PnP} and BPnPNet \cite{blind-pnp}. Diff. PnP estimates the camera pose from the predicted correspondences. However, pose accuracy is highly sensitive to correspondence quality, making the pose loss less reliable during training. Although the declare network \cite{Declarative_network} in BPnPNet \cite{blind-pnp} is an effective module in optimizing blind PnP, it requires an accurate pose prior. In BPnPNet \cite{blind-pnp}, the pose loss computed from the filtered correspondences has a limited impact on the I2P registration architecture. The proposed MinCD-Net detects and learns 2D-3D keypoints uniformly distributed in the 2D and 3D spaces, which achieves a higher learning efficiency and is more robust to correspondence quality.

\begin{table}[t]
\scriptsize
\centering
\caption{Recall and precision of the learned 3D keypoints. Precision and recall are computed with respect to the pre-detected 2D keypoints (pixel threshold is $3$). Avg. Num represents the average number of learned 3D keypoints.}
\begin{tabular}{c|ccccc}
\hline
\textbf{Parameter} $s_{\text{th}}$ & $e^{-0.1}$ & $e^{-0.2}$ & $e^{-0.3}$ & $e^{-0.4}$ & $e^{-0.5}$ \\
\hline
\textbf{Precision} & N/A & 0.582 & 0.531 & 0.442 & 0.308 \\
\textbf{Recall}    & N/A & 0.454 & 0.562 & 0.722 & 0.862 \\
\textbf{Avg. Num}    & N/A & $\approx$3.1K & $\approx$5.7K & $\approx$8.4K & $\approx$14.2K \\
\hline
\end{tabular}
\label{tab:ab_1}
\end{table}

\begin{table}[t]
\scriptsize
\centering
\caption{Ablation study of the different learning schemes. Model was trained on the Office scene and tested on the remaining scenes.}
\begin{tabular}{c|ccc|c}
\hline
\textbf{Schemes} & $L_{\text{corr}}$  & $L_{\text{corr}}+L_{\text{key}}$ & $L_{\text{corr}}+L_{\text{key}}+L_{\text{Chamfer}}$ & \textbf{Gain} \\
\hline
\textbf{IR} & 0.473 & 0.489  & \textcolor{blue}{\textbf{0.567}} & $\uparrow$ \textbf{9.4}\% \\
\textbf{RR} & 0.502 & 0.516  & \textcolor{blue}{\textbf{0.646}} & $\uparrow$ \textbf{14.4}\% \\
\hline
\end{tabular}
\label{tab:ab_2}
\end{table}

\subsection{Ablation studies}

To investigate the performance of MinCD-Net, we conduct the ablation studies of the hyper-parameter $s_{\text{th}}$ and the loss functions. We analyze the relationship between $s_{\text{th}}$ and the quality of learned 3D keypoints. As presented in Table \ref{tab:ab_1}, if $s_{\text{th}} \geq e^{-0.1}$, no 3D keypoints are retained. If $s_{\text{th}}$ is set too low, a large number of redundant 3D keypoints are learned that disturbs Chamfer distance minimization. To balance precision and recall, $s_{\text{th}}$ is best set to $e^{-0.4}$, and the visualization of the learned 3D keypoints is provided in Fig. \ref{fig:vis-2}. With the fixed optimal $s_{\text{th}}$, MinCD-Net ranks $1st$ on four datasets. It suggests that $s_{\text{th}}$ tuned in one dataset has stable and accurate performance in other datasets.

Then, we study the different loss functions in Table \ref{tab:ab_2}. It is unsurprising that the loss $L_{\text{corr}}+L_{\text{key}}$ shows only a minor improvement over $L_{\text{corr}}$, as $L_{\text{key}}$ supervises only 3D keypoints, which are not incorporated into the network's main branch. $L_{\text{Chamfer}}$ plays a dominant role, as it acts as a global geometrical constraint. 

We also investigate the dependency of MinCD-Net on 2D keypoint detectors, like FAST \cite{fast}, SIFT \cite{sift}, Superpoint \cite{superglue}, and even the uniformly sampled scheme. Results in Table \ref{tab:ab_3} indicate that MinCD-Net achieves nearly the same results with other common detectors, even the uniformly sampling. So, MinCD-Net needs  detected 2D keypoints, but not relies on the specific detector. Besides, the computation analysis of the current methods are provided in Table \ref{tab:ab_4}. It indicates that MinCD-Net is a lightweight network with few extra runtime and GPU memory. Overall, the above results show the effectiveness of MinCD-Net.

\begin{table}[t]
\scriptsize
\centering
\caption{Ablation study of the proposed method with the different choice of 2D keypoint detectors.}
\begin{tabular}{c|ccccc}
\hline
\textbf{Schemes} & Shi-Tomasi & FAST & SIFT & SuperPoint & Uni. sampled \\
\hline
\textbf{IR} & 0.567 & 0.552 & 0.572 & 0.560 & 0.542 \\
\textbf{RR} & 0.646 & 0.631 & 0.638 & 0.649 & 0.625 \\
\hline
\end{tabular}
\label{tab:ab_3}
\end{table}

\begin{table}[t]
\scriptsize
\centering
\caption{Computation efficiency analysis of the current methods. Diff. PnP, BPnPNet, and MinCD-Net are only used to supervise the backbone networks (not used in the inference stage), so that we record runtime and GPU memory in the training stage.}
\begin{tabular}{c|ccc|c}
\hline
\textbf{Methods} & Runtime/ms & Param/M & GPU memory/MB & RR \\
\hline
Baseline & 127 & 28.2 & 7532  & 51.0\% \\
+Diff. PnP & 152 (+25) & 28.2 (+0.0) & 7852 (+320) & 49.1\% \\ 
+BPnPNet   & 141 (+14) & 30.8 (+2.6) & 8242 (+710) & 57.8\% \\ 
 +\textbf{MinCD-Net} & 148 (+21) & 31.4 (+3.2) & 8353 (+821) & \textbf{64.7}\%  \\
\hline
\end{tabular}
\label{tab:ab_4}
\end{table}





\section{Conclusions}

To achieve more accurate I2P registration, we leverage the blind PnP into correspondence learning. First, we simplify blind PnP to a more amenable task MinCD-PnP. It ensures the feasibility of learning correspondences with blind PnP. To effectively solve MinCD-PnP, we develop a lightweight multi-task learning module, MinCD-Net. It can be easily integrated into the I2P registration networks. Extensive experiments on four indoor datasets demonstrate that MinCD-Net achieves a superior IR and RR metrics compared to the existing I2P registration methods in both cross-scene and cross-dataset setting. 

\noindent \textbf{Limitations and future work}. In the challenging scenarios (i.e., self-collected dataset), the gain of MinCD-Net is not substantial (as seen in Table \ref{tab:res_2}). The precision of learned 3D keypoints is not high (as seen in Table \ref{tab:ab_1}). To address these limitations, we plan to use the learnable correspondences pruning module \cite{an-yang-add-1} to improve the solving efficiency of MinCD-PnP task.

\section*{Acknowledgments}

This work is supported by the National Key R\&D Program of China (2024YFC3015303), National Nature Science Foundation of China (62372377) and China Postdoctoral Science Foundation (2024M761014).

{
    \small
    \bibliographystyle{ieeenat_fullname}
    \bibliography{main}
}

\end{document}